\pdfoutput=1

\documentclass[11pt]{article}

\usepackage[]{EACL2023}

\usepackage{times}
\usepackage{latexsym}
\usepackage{graphics}
\usepackage{graphicx}
\usepackage{hypcap} 
\usepackage{amsfonts}

\usepackage[T1]{fontenc}

\usepackage[utf8]{inputenc}

\usepackage{microtype}

\usepackage{inconsolata}

\title{Exploring Category Structure with Contextual Language Models and Lexical Semantic Networks}

\author{Joseph Renner$^1$   Pascal Denis$^1$   Rémi Gilleron$^1$  Angèle Brunellière$^2$ \\ \\
    $^1$Univ. Lille, Inria, CNRS, Centrale Lille, UMR 9189 - CRIStAL, F-59000, Lille, France \\ 
            \texttt{\{firstname.lastname\}@inria.fr} \\
    $^2$Univ. Lille, CNRS, UMR 9193 - SCALab, F-59000 Lille, France\\
                    \texttt{angele.brunelliere@univ-lille.fr}     
}

\begin{document}
\maketitle
\begin{abstract}
Recent work on predicting category structure with distributional models, using either static word embeddings \cite{heyman2019can} or contextualized language models (CLMs) \cite{misra_et_al}, report low correlations with human ratings, thus calling into question their plausibility as models of human semantic memory. In this work, we revisit this question testing a wider array of methods for probing CLMs for predicting typicality scores. Our experiments, using BERT \cite{devlin2018bert}, show the importance of using the right type of CLM probes, as our best BERT-based typicality prediction methods substantially improve over previous works. Second, our results highlight the importance of polysemy in this task: our best results are obtained when using a disambiguation mechanism. Finally, additional experiments reveal that Information Content-based WordNet \cite{wordnet}, also endowed with disambiguation, match the performance of the best BERT-based method, and in fact capture complementary information, which can be combined with BERT to achieve enhanced typicality predictions. 

\end{abstract}

\section{Introduction}
\label{intro}
The empirical success of contextual language models (CLMs) \citep{ELMO, devlin2018bert, liu2019roberta} has led to much research analyzing their functionality \citep{bertology, wu2020perturbed, ethayarajh2019} and how they acquire semantic and world knowledge \citep{petroni2019}. It also raises the question of their plausibility as models of human semantic memory \cite{chronis2020bishop,ettinger2020bert,gari-soler-apidianaki-2021-lets}. The study of categorical knowledge, in particular \textit{typicality}, provides a window into this question \cite{murphy2004big}. As initially observed by \citet{rosch1975cognitive}, native speakers of English consider that certain exemplars (e.g., \texttt{robin}, \texttt{crow}) are more representative than others (e.g., \texttt{penguin}, \texttt{ostrich}) of a conceptual category (\texttt{birds}). That is, categorical knowledge is organized along a graded structure. According to prototype theory \citep{rosch1975cognitive}, the structure follows from the fact that properties frequently shared among category members tend to be integrated into its prototype.  

The main question we raise in this paper is whether distributional models, and CLMs in particular, are indeed aware of category structure, as captured in existing human typicality norms for English. The broader question for cognitive science is whether or not this type of knowledge can be learned through text-based exposure alone, thus contributing to the larger debate on the role of language in learning semantic knowledge \cite{lupyan2019words}. But there are also more practical motivations for this work, as many NLP tasks (such as information retrieval, question answering, natural language inference) can arguably benefit from typicality relations by understanding which exemplars are more relevant to particular concepts. 

Previous work on this topic provide mostly negative results. \citet{heyman2019can} use classical static embeddings and represent typicality scores between categories and their exemplars as the cosine distance between their corresponding word embeddings. They conclude that static embeddings poorly account for human typicality scores, as obtained from \citet{morrow2005representation}. More recently, \citet{misra_et_al} use various CLMs to predict typicality. While CLMs have a larger, more expressive parameter space and are tuned over larger corpora with arguably better learning objectives than static word embeddings, \citet{misra_et_al} report only slightly better, and still modest, correlations with human typicality judgements (in this case, \citet{rosch1975cognitive}). Their probing approach uses a Cloze test formulation over taxonomic sentences (e.g., \textit{A robin is a bird}). 

We revisit this question by first introducing an expanded suite of methods to extract typicality scores from CLMs, some of which improve correlations with human norms, showing the importance of using the right probe for typicality. How to reliably and efficiently probe CLMs is still an open question, including supervised and unsupervised approaches \cite{bertology,elazar-etal-2021-amnesic,wu2020perturbed}. The problem is exacerbated by the fact that typicality is a relation between pairs of concepts, abstracted away from contexts, while CLMs produce representations for contextualized word tokens (not types). We consider a wider array of approaches, all unsupervised, for predicting typicality from CLMs, using BERT as representative test case.  Our probing approaches fall into two main classes: (i) BERT as a probabilistic language model, and (ii) generating "static" word embeddings from contextual BERT embeddings. 

Our second contribution is showing the importance of polysemy (e.g., \textit{orange} can be a fruit, a color, or a company) for typicality, an issue that has been largely overlooked in previous works. When judging category-exemplar pair typicality, humans arguably don't consider all the senses of the category and exemplar words. Instead they consider the senses of the category and exemplar that are most compatible with one another, in effect performing a joint disambiguation. Polysemy is problematic for static word embeddings as they collapse word senses into a single vector. CLMs, on the other hand, provide some form of sense selection through contextualization \citep{ethayarajh2019,gari-soler-apidianaki-2021-lets}, but there are many possible ways to use CLMs and to provide contexts to these models (such as sampled or taxonomic sentences). We find that using a simple disambiguation mechanism, such as the method of deriving multi-prototype embeddings introduced in \citet{chronis2020bishop}, leads to typicality predictions more closely correlated with human rankings.

Finally, another research question we address in this paper is whether categorical structure is present in lexical semantic networks, such as WordNet \cite{wordnet}. The hierarchical concept organization in WordNet was indeed initially inspired by theories of human semantic memory \cite{beckwith2021wordnet}. Note that polysemy is also an issue in using WordNet, as we need to map category and exemplars to specific synsets. However, using the structure of WordNet, a simple disambiguation heuristic, and taking into account word frequency information leads to competitive results. Furthermore, we find that WordNet-based and BERT-based typicality predictions contain complementary information, and creating a simple ensemble method of the two further increases performance.

In summary, our main contributions are:


\begin{itemize}
    \itemsep-3pt 
    \item We introduce and compare a large array of unsupervised probing approaches for assessing whether BERT capture the internal structure of semantic categories. 
    \item Our experiments reveal BERT can predict human typicality rankings more reliably than previously found in \citet{heyman2019can} and \citet{misra_et_al}, but only when endowed with a disambiguation mechanism.
    \item Similarly, typicality predictions based on the structure of WordNet (again with a disambiguation mechanism) can achieve similar but complementary performance.
    \item A simple combination of BERT and WordNet predictions leads to a higher level of correlation (Spearman greater than 0.5) with human typicality rankings. 
\end{itemize}

\section{Related Work}
\label{related}

Recent years have seen an ever growing body of work on assessing whether distributional models constitute realistic models of human semantic memory, within both NLP \cite{chronis2020bishop,ettinger2020bert} and Cognitive Science \cite{hollis2017estimating, hollis2016principals, mandera2017explaining,gunther2019vector,lupyan2019words,kumar2021semantic}. In this context, the study of categorical knowledge, and specifically graded typicality, plays a crucial role, given it is one of the most reliable findings in the study of human categorical knowledge \cite{rosch1975cognitive,murphy2004big}. The first work on using distributional models to predict typicality scores is \citet{heyman2019can}, who estimate these scores, in English and Dutch, using cosine distances between the static embeddings of the exemplar and category words. These show that static embeddings, whether trained in a counted- or prediction-based fashion, yield poor correlations with human typicality norms. On the related task of lexical entailment (LE), \citet{vulic2017hyperlex} evaluated different unsupervised and supervised methods that use static word embeddings. They also found a significant gap between static word embeddings and human norms. While LE is closely related to typicality, it differs in that LE predicts the relationship between two words, while typicality compares the relationship between two concepts (in which words are used as a proxy, in our case). Also, LE ratings are not restricted to pairs of words belonging to the same category. We differ from these earlier work on typicality by considering more expressive CLMs instead of static embeddings. The most closely related work is \citet{misra_et_al} who probe a wide range of CLMs (i.e., different variants of BERT and GPT) for predicting typicality scores: these are extracted using a Cloze task formulation over hand crafted taxonomic sentences (e.g. \textit{A robin is a bird.}). They report better correlation scores with human norms, although still modest, which they take as indication that text exposure is not sufficient to learn category structure. Our focus in this work is different and somewhat broader in that we consider a wider array of probing methods of CLMs, though restricted to unsupervised ones. Supervised probing approaches, based on classifiers taking CLM representations as inputs, are problematic as they provide a less direct probing approach, possibly adding additional confounds \citep{elazar-etal-2021-amnesic, wu2020perturbed}. Another distinctive aspect of our work is that we compare CLMs to lexical networks like WordNet \cite{wordnet}. Furthermore, we study the impact of polysemy in typicality, which was not controlled in \citet{misra_et_al} or in \citet{heyman2019can}. The work of \citet{apidianaki-gari-soler-2021-dolphins} shares some similarities with \cite{misra_et_al}: they also use BERT under a Cloze task setting to study typicality, but they focus on identifying whether the model has access to prototypical properties of concepts (e.g., a ball is round).



\section{Problem Setting and Framework}

\label{problem_setting}


This section presents our general framework for assessing whether a lexical representation model (i.e., a CLM or semantic network) is aware of category structure; the specific probing approaches will be described in Sec.~\ref{sec:methods}. Let us assume a generic conceptual space consisting of a set $\mathcal{C}$  of categories as well as a set $\mathcal{E}$ of exemplars of these categories. For each category $\mathtt{c}$ in $\mathcal{C}$, we assume a set of $n_c$ exemplars $\mathtt{e}_i^\mathtt{c}$ in $\mathcal{E}$ for $i=1$ to $n_c$. Following the findings of \citet{rosch1975cognitive}, we posit that this conceptual space has a graded structure, in that certain exemplars (e.g., \texttt{robin}, \texttt{crow}) are more closely representative of their category (e.g., \texttt{birds}) than others (e.g., \texttt{penguin}, \texttt{ostrich}). This can be expressed through a typicality score $t_i^c \in \mathbb{R}^{+}_0$, often an ordinal scale, for each $\mathtt{e}^\mathtt{c}_i$, where $t_i^c > t_j^c$ indicates that exemplar $\mathtt{e}^\mathtt{c}_i$ is more typical of category $\mathtt{c}$ than $\mathtt{e}^\mathtt{c}_j$. These typicality scores can be obtained from human subjects through various kinds of stimuli, such as direct scoring of category-exemplar word pairs (e.g., \textit{robin}-\textit{bird}) and scoring of a taxonomic sentence (e.g., \textit{A robin is a bird.})(see Sec.~\ref{sec:datasets}).

To assess whether a lexical representation model is aware of such a structure, three main components are needed. As the conceptual space is latent, we first need a \textbf{concept-to-word mapping function} from category and exemplar concepts to words. As is done in human studies and previous work, we assume that categories and exemplars are reliably accessed through their corresponding singular words: the concepts \texttt{robin} and \texttt{bird} are accessed via the words \textit{robin} and \textit{bird}, respectively. This is a simplification, as the same concept can be realized by plural forms, as well as by different (synonymous) words. Given this functional mapping, we will conflate a category $\mathtt{c}$ with its category word $c$ and an exemplar $\mathtt{e}^\mathtt{c}_i$ with its exemplar word $e^c_i$.

Second, we define a \textbf{typicality scoring function} over exemplar-category word pairs, which captures how typical the exemplar is of the category. Each model will assign a typicality score $s_i^c$ to each category $c$ and exemplar $e_i^c$ pair. Some of our methods will make use of a text corpus, denoted $S$, taking the form of a pre-extracted \textbf{set of sentences} containing exemplar and/or category words. This corpus $S$ can be viewed as an extra parameter of the typicality scoring function, as it will directly impact the predicted typicality values $s_i^c$. Such a corpus is required to be able to fully leverage CLMs, whether they are used as probabilistic language models or to produce word vector representations. It is also through this corpus that one hopes to capture contextual word realizations that approximate the different senses associated with the exemplar and category words. Indeed, we hypothesize that one important aspect of assessing typicality between category and exemplar words is to be able to deal with the \textbf{polysemy} of these words, predicting that the best methods should be able to disambiguate these words in such a way that the exemplar word is interpreted in a category-compatible sense and the category word in an exemplar-compatible sense. Using WordNet synsets as a proxy for senses, we find that on average, category words are linked to $4.2$ and $4.9$ senses, while exemplars are linked to $3.3$ and $4.7$ senses on the typicality datasets of \citet{morrow2005representation} and \citet{rosch1975cognitive}, respectively\footnote{If restricting synsets to only nouns, the average number of category senses $2.9$ and $3.4$ for the two respective datasets, and $2.4$ and $3.1$ for exemplars.} (see Sec.~\ref{sec:datasets} for dataset details). This shows the depth of the polysemy problem for typicality, as each word on average has multiple meanings.

Finally, we need an \textbf{evaluation metric} for measuring how well the predicted typicality scores $s_i^c$ are able to mirror the human judgements $t_i^c$. The most obvious choice for this metric is to use Spearman rank correlation \citep{spearman1904}, as this correlation measure is non-parametric and makes no strong (e.g., linear) assumption on the data distribution or the underlying scoring functions.

\section{Methods}
\label{sec:methods}
We distinguish two classes of methods, depending on whether they use CLMs or WordNet. For CLM-based methods, we use BERT (\texttt{bert-base-uncased} specifically) as a prototypical CLM to assess whether CLMs are aware of category structure. The framework and methods easily generalize to other CLMs. While the concept-to-word mapping is the same for these methods, the scoring function and the probing corpus are different. The BERT methods fall into two categories: (i) those that use BERT as a language model (BERT-MLM, BERT-SentEmb, BERT-MLM-Taxo), and (ii) those that extract word vectors from BERT (BERT-Avg, BERT-MPro). Note that these are all unsupervised probing methods: when an additional corpus $S$ is used, it is only used to probe BERT, not to fine-tune it. These methods embody different linguistic hypotheses and adopt different ways of dealing with polysemy. Note that for embedding methods, we test all layers.

Secondly, we consider lexical semantic networks to compute typicality scores, using WordNet \cite{wordnet} specifically. For this, we use the Shortest-Path and the Lin \cite{lin1998information} similarity measures computed on WordNet \citep{wordnet}. We compute the information content (IC) values \citep{resnik1995using} used in the Lin measure using the additional corpus $S$. We deal with polysemy by using the maximum similarity between exemplar and category synset pairs to compute typicality scores.

\subsection{Using BERT's Language Model} 
\label{BERT:pretrain}
The first class of methods relies on the language modeling abilities of BERT, using the following hypothesis: the more central (or peripheral) an exemplar is to a category, the more (or less) likely it is to be used in the category context(s). This hypothesis can be turned into two different distributional hypotheses, depending on whether we take a \textit{paradigmatic} perspective (i.e., how likely can the exemplar word be substituted for the category word in the category contexts) or a \textit{syntagmatic} perspective (i.e., how likely can the exemplar be used next to its category). 
Our first two probing methods take a paradigmatic perspective and are reminiscent of the Distributional Inclusion Hypothesis, first proposed by \citet{geffet2005distributional}. They compute typicality scores that reflect how successful the substitution of a category word by an exemplar is, measured by conditional word probability (BERT-MLM) or by cosine distance between sentence embeddings before and after substitution (BERT-SentEmb). In both, the set of sentences is restricted to sentences that contain the category words and is denoted by $S_c$. 

\subsubsection{Masked Language Modeling (BERT-MLM)} 
Under this approach, typicality is computed as the conditional probability of seeing the exemplar word in its category's contexts. Specifically, for each sentence in $S_c$, the category name is masked using the \texttt{[MASK]} token and the resulting sentence is passed through BERT and the MLM classification, yielding MLM logits for each subtoken in the vocabulary, which are softmaxed to probabilities. Typicality scores are obtained by averaging across sentences the MLM probabilities for each exemplar subtoken sequence. Formally, the typicality score $s_i^c$ for an exemplar $e_i^c$ with $l_i$ subtokens and category $c$, given BERT and masked sentences $S_c$, is computed by

\begin{equation} \label{eq:MLM}
s_i^c = \frac{1}{|S_c|}\sum_{j=1}^{|S_c|} \prod_{k=1}^{l_i} BERT_{MLM}(S_{c_k}^j)
\end{equation}

\noindent where $S_{c_k}^j$ is the $k$th subtoken of exemplar $e_i^c$ in the $j$th sentence of $S_c$, and $BERT_{MLM}(\cdot)$ gives the probability of this subtoken.

\subsubsection{Sentence Embedding Modeling (BERT-SentEmb)}
In this method, typicality is taken to be a measure of how well a category word can be substituted by its exemplars in sentences $S_c$ without altering the overall sentence meaning, which we approximate as the sentence embedding. Recall that in BERT's NSP objective, the \texttt{[CLS]} sentence classification token is input to the NSP classification head. For each category name sentence, we compute layer wise activations of this sentence token. Then, for each exemplar, we replace the category word in each sentence with the exemplar and obtain the same activations. The typicality score is the cosine similarity of the original sentence embedding and that of the replaced sentence, averaged across sentences. The sentence embeddings can be obtained from any layer in the model; we test the method separately for all layers. Formally, given a category $c$ and an examplar $e_i^c$, a set of sentences $S_c$ containing the category word $c$, and a $SE(\cdot)$ function which outputs a sentence embedding from a specific layer, the typicality score is computed by
\begin{equation}
s_i^c = \frac{1}{|S_c|}\sum_{j=1}^{|S_c|} cos(SE(S_{c}), SE(S^j_{c\rightarrow e_i^{c}}))
\end{equation}
\noindent where $S^j_{c\rightarrow e_i^c}$ is $S^j$ with $c$ replaced by $e_i^c$, and $cos(\cdot,\cdot)$ is the cosine similarity operator. 

As they rely on paradigmatic substitution, neither BERT-MLM nor BERT-SentEmb explicitly attempt to disambiguate exemplar or category words, but one can argue that some form of sense selection happens through BERT's contextual modeling and the selection of category contexts. This ensures that the word \textit{orange} is used in \textit{fruit} compatible contexts, thus hopefully filtering out contexts compatible with other senses (e.g. color), when predicting typicality within the \texttt{fruit} category. But there is no disambiguation of the category word, as all of its contexts are randomly sampled. This might introduce some noise as category examples with another sense (e.g., \textit{fruit of their labor}) or even POS (e.g., \textit{the trees fruit early}) might be sampled.

\subsubsection{Masked Language Modeling with Taxonomic Sentences (BERT-MLM-Taxo)}
\label{MLM-taxo}
Inspired by \citet{misra_et_al}'s Taxonomic Sentence Verification method, our third CLM-based method is similar to BERT-MLM, but different in that it uses \textit{taxonomic} propositions instead of sampled sentences. These propositions are in the form \textit{"A(n) X is a(n) Y"}, where \textit{X} and \textit{Y} are exemplars $e_i^c$ and categories $c$, respectively. Typicality scores are obtained in the same way as BERT-MLM: exemplar subtokens are masked in the taxonomic proposition, MLM logits are obtained from the model, then the product of the softmaxed logits (probabilities) is the typicality score (same as Eq.~\ref{eq:MLM}). While these sentences provide only narrow contexts, and are somewhat artificial and different from the BERT's training data, they are informative in providing implicit mutual disambiguation of \textit{both} the exemplar and category words. 

BERT-MLM-Taxo is similar to \citet{misra_et_al}, but it differs in two ways. First, \citet{misra_et_al}'s method uses CLMs to compute the probability of the \textit{category} word in the taxonomic proposition (i.e. $P(c|e_i ^c)$), while our method does the opposite ($P(e_i^c|c)$) to more resemble how humans are probed for the task. Second, \citet{misra_et_al} calculates conditional probabilities for masked language models such as BERT using the formulation introduced in \citet{bert_mrf}: separately masking each subtoken in the word and passing these separately masked variants of the taxonomic sentence through the model, summing the masked subtokens across variants. While this method is more theoretically founded (as BERT is not an incremental language model and cannot compute probabilities through the chain rule), it is subject to skewed MLM logits if used on exemplars\footnote{All category words are treated as one subtoken.}, as a result of words being broken into subtokens (example from \citet{misra_et_al}: if the word ostrich is segmented into \textit{ostr} and \textit{ich}, then the probability of \textit{ich} given that it is preceded by \textit{ostr} is anomalously high, skewing the sequence probability). Thus, we mask all subtokens of the exemplar and treat the resulting probabilities as a sequence.

\subsection{BERT-Based "Static" Representations} 
\label{BERT:static}

The next CLM-based methods derive static word representations from contextual representations computed over a corpus $S$ which includes sentences containing category words as well as sentences containing exemplar words. We use cosine similarity between category and exemplar embeddings for typicality scores $s_i^c = cos(R(e_i^c), R(c))$, where $R(\cdot)$ provides the static representations of the category and exemplar. We derive static word representations by averaging all hidden state activations of a word over $S$ (BERT-Avg), or clustering these hidden states, allowing for sense modulation (BERT-MPro). For this set of methods, our hypothesis is that the vectorial space induced by BERT captures typicality, as opposed to the language modeling capabilities used in the methods of Sec.~\ref{BERT:pretrain}.

\subsubsection{Averaged Contextual Embeddings (BERT-Avg)}
The first approach for generating static word \textit{type} vectors from contextualized word \textit{token} embeddings is to simply average them \citep{bommasani-etal-2020-interpreting}. For each sentence in $S$, we compute and store the contextual representations of the word at each layer of the BERT model. We then average these representations over sentences for each word, giving a static embedding for each layer, for each category and exemplar. The typicality score of a pair $(e_i^c,c)$ is computed as the cosine similarity between the static embeddings of the representation of $e_i^c$ and of the representation of $c$. As with BERT-SentEmb, we test the method separately for all layers. Note that this approach inherits the problem found in classical static embeddings that it conflates all possible senses of a word. 

\subsubsection{Multi-Prototype Contextual Embeddings (BERT-MPro)}
To further exploit context, we use multi-prototype BERT embeddings \citep{chronis2020bishop}. For each category name and exemplar, we use $k$-means to cluster the word's contextual embeddings computed from its sampled sentences, yielding, for a given $k$, $k$ cluster centroids for each layer, which disambiguate the $k$ different possible meanings of the word. Following \citet{chronis2020bishop}, we predict scores using the $maxsim(\cdot,\cdot)$ function between the cluster centroids of the category name and exemplar, which yields the maximum similarity value of category-exemplar centroid pairings. Formally, for a given $k$ and layer $l$, and set of cluster centroids $\tau(w)_1^l ...  \tau(w)_k^l$ for category $c$ and exemplar $e_i^c$, the similarity $s_i^c$ is defined by

\begin{equation}
maxsim(c, e_i^c) = \max\limits_{1 \leq j \leq k, 1 \leq t \leq k} cos(\tau(c)_j^l, \tau(e_i^c)_k^l)
\end{equation}

\noindent where $cos(\cdot,\cdot)$ is the cosine similarity measure. Following \citet{chronis2020bishop}, we sidestep tuning the number of clusters $k$ and simply take as input to the $maxsim$ operator the union of all clusters from $k\leq15$. Furthermore, we also test the embeddings from all layers separately.

\subsection{WordNet-Based Methods}
Our second class of methods uses WordNet \citep{wordnet} to compute typicality scores. Let us suppose that a category word $c$ and exemplar word $e_i^c$ are mapped to synsets $s(c)$ and $s(e_i^c)$, respectively. Only noun synsets are considered for $c$ and $e_i^c$. The hypothesis here is that the more closely linked $s(c)$ and $s(e_i^c)$ are in the WordNet graph, the more the exemplar is typical of the category. To measure this "linkage" between exemplars and categories, we use two WordNet similarity measures, thus yielding two sets of WordNet-based methods. 

\subsubsection{Shortest Path (WNSP, WNSP-noWSD)} 
Our first similarity is the shortest path between $s(c)$ and $s(e_i^c)$ along hypernym/hyponym edges of WordNet. Given that exemplars and category words might be linked to multiple synsets, we need to aggregate the similarities between synsets. A first method, called \textbf{WNSP-noWSD}, is to average the similarities between all synsets $s(c)$ and $s(e_i^c)$. The second method (\textbf{WNSP}) tries to disambiguate between possible synsets, relying on the $maxsim(\cdot,\cdot)$ operator, but defined over synset similarities (instead of cosine similarity over clusters, as in BERT-MPro). Formally, for a category name $c$  and exemplar $e_i^c$, the typicality score $s_i^c$ is equated to $maxsim(c, e_i^c)$, which is defined by  

\begin{equation} \label{eq:WNsim}
maxsim(c, e_i^c) = \max\limits_{s(c), s(e_i^c)} sim(s(c), s(e_i^c)) 
\end{equation}

\noindent where $sim(\cdot)$ denotes the shortest path similarity. 

Note that, irrespectively of the disambiguation process, the shortest path similarity possibly lacks expressivity, as synsets of different exemplars can be at the same distance of the category synset. 

\subsubsection{Lin Similarity (WNIC, WNIC-noWSD)} 
For a more expressive method, we use the Lin similarity measure \cite{lin1998information}, which computes similarity using the most specific ancestor node and Information Content (IC) (computed via Wikipedia text dump), a measure of specificity for a concept closely linked to frequency \cite{resnik1995using}, thus combining frequency with hierarchical semantic knowledge. \citet{pedersen2010} show that augmenting WordNet with IC results in higher correlation with human judgments on similarity and relatedness tasks. Formally, let  $IC(\cdot)$ and $lcs(\cdot,\cdot)$ denote information content and least common subsumer (specific ancestor node), respectively, the similarity between synsets $s(c)$ and $s(e_i^c)$ is defined by

\begin{equation}
lin(s(c), s(e_i^c)) = \frac{2 * IC(lcs(s(c), s(e_i^c)))}{IC(s(c)) + IC(s(e_i^c))}
\end{equation}

\noindent We then can aggregate similarities by averaging, which defines the method \textbf{WNIC-noWSD}, or by using Equation~\ref{eq:WNsim} where $sim(\cdot)$ is chosen to be the Lin similarity. This latter method is called \textbf{WNIC}.

\section{Experiments}
\label{experiments}

\subsection{Datasets}
\label{sec:datasets}
As ground truth typicality ratings, we use three datasets of human typicality ratings, one from \citet{rosch1975cognitive} (young adult ratings) and two from \citet{morrow2005representation} (young and older adult ratings). The two dataset sources were produced at different time periods (1975 for Rosch and 2005 for M\&D), with different sample sizes ($209$ participants and 54, respectively) and potential cultural differences (US residents and UK residents, respectively) and with different protocols. Human ratings were obtained by scoring a taxonomic sentence for Rosch and by scoring category-exemplar pairs for M\&D. The datasets contain varying number of categories (11 for M\&D, 10 for Rosch) and exemplars per category ($29$-$128$ for M\&D, $42$-$54$ for Rosch), with $7$ overlapping categories between the two sources.\footnote{See Table \ref{table:dataset} in Appendix for category/exemplar statistics.} The human ratings between the two splits of M\&D are strongly correlated (average category Spearman correlation: $0.857$, see Appendix \ref{dataset_agreement}). However, the correlation between the Rosch rankings and M\&D rankings on the category-exemplar intersection of the datasets are much lower, likely because of the differences outlined above (average category Spearman: $0.656$ and $0.574$ for Rosch vs M\&D young and older adults, respectively). We follow the preprocessing procedure of \citet{heyman2019can}, (see Appendix \ref{dataset:preprocessing}). Our auxiliary corpus $S$ is extracted by sampling sentences from Wikipedia, described in Appendix~\ref{dataset:sent_sampling}.

\subsection{Baseline and Competing Methods}
\label{baselines}

\paragraph{Word Frequency}
The first baseline is typicality scores as the exemplar's number of occurrences in the corpus $S$. It is natural to think that more typical examples are more popular words, and \citet{heyman2019can} show that frequency partly accounts for human ratings in the M\&D datasets.

\paragraph{PPMI-SVD, Word2Vec}
Other baselines are the cosine similarities between the category-exemplar static word embeddings from count-based (singular value decomposition of positive pointwise mutual information matrix: \textbf{PPMI-SVD}) and prediction-based (\textbf{Word2Vec} skip-gram negative sampling) algorithms. For baseline details, see Appendix \ref{baseline_details}.

\paragraph{Taxonomic Sentence Verification}
We use the Taxonomic Sentence Verification method of \citet{misra_et_al}, using \texttt{bert-base-uncased}. The differences with BERT-MLM-Taxo can be found in  Sec.~\ref{MLM-taxo}.

\subsection{Results and Analysis}
\label{results_and_discussion}
\subsubsection{Method Performance}
Mean Spearman correlations for all methods are reported in Table~\ref{table:all-results}. \footnote{See Table~\ref{table:category-results} in Appendix~\ref{app:category} for results by category.} 

\paragraph{BERT-based Methods}
Our first main result is that the BERT-based methods\footnote{Recall that we use \texttt{bert-base-uncased}.} all improve over baselines on the M\&D datasets. For the Rosch dataset, only BERT-MPro and \citet{misra_et_al}'s method improve over Word2Vec. BERT-MPro has the highest correlations of the BERT based methods across datasets, with \citet{misra_et_al}'s method only slightly edging it on the Rosch dataset.

Comparing the substitution-based methods from Sec.~\ref{BERT:pretrain}, BERT-SentEmb and MLM-Taxo have similar performance. BERT-MLM-Taxo performs better than BERT-MLM, showing that taxonomic sentences provide narrow yet more informative contexts than a randomly sampled corpus simply based on the category word, as they allow for mutual disambiguation of category and exemplar words. \citet{misra_et_al}'s method performs worse than BERT-MLM-Taxo on the M\&D dataset but better on the Rosch dataset.\footnote{This was the only dataset used in \citet{misra_et_al}.} These discrepancies across datasets can be attributed to different protocols for obtaining human ratings: scoring category-exemplar pairs for M\&D and scoring a taxonomic sentence for Rosch. They also account for the performance differences of the frequency baseline, which yields higher correlation scores on M\&D than on the Rosch dataset. Typicality scores are dissociable from lexical frequency when subjects judge whether a sentence is a good example of their idea of the category. Our results are in line with \citet{Mervis1976}, showing that Rosch dataset scores are not correlated with frequency. 

Comparing the word embedding methods from Sec.~\ref{BERT:static}, the performance increase from BERT-Avg to BERT-MPro shows the importance of disambiguation. Layer-wise performance can be found in Appendix~\ref{sec:avg_MP_hyper_perf}. It should be noted that for BERT-MPro, the best performing layer was layer 10, confirming \citet{chronis2020bishop}'s claim that similarity-based task performance peaks in layers 8-10,\footnote{They denote this as layers 7-9 in their paper (0-indexed); we index starting at 1 to denote the embedding layer as 0.}. Later layers show more contextual variation \cite{ethayarajh2019} capturing finer-grained sense differences. Inversely, for BERT-Avg earlier layers perform best as they include less contextual variation, making the mean more stable. 

\paragraph{WordNet-based Methods}

For the WordNet-based methods, WNIC achieves the highest correlations across datasets, and is competitive with the best BERT-based methods. This confirms the findings of \citet{pedersen2010} on other intrinsic tasks such as similarity. Disambiguation in the WordNet methods seems crucial, as the variants using averaging perform worse than their $maxsim$ versions. WNIC's high performance shows the importance of frequency information. 

\paragraph{Conclusion} The experimental results confirm the importance of handling polysemy for both BERT- and WordNet-based methods. Note that the best performing method (last row), which combines BERT-MPro and WNIC, is discussed in Sec.~\ref{sec:combining}.

\begin{table*}[h]
\centering
\small
\resizebox{3.3in}{!}{
\begin{tabular}
{ | l | l | c | c | c |} \hline
Method
Class & Method & M\&D-Young & M\&D-Old & Rosch \\ \hline
Baselines & Frequency & 0.323 & 0.296 & 0.059  \\
    & PPMI-SVD & 0.236 & 0.227 & 0.294 \\
    & W2V & 0.260 & 0.296 & 0.338 \\ \hline
BERT & BERT-SentEmb & 0.381 & 0.373 & 0.289 \\
    & BERT-MLM & 0.354 & 0.358 & 0.238 \\
    & BERT-MLM-Taxo & 0.396 & 0.393 & 0.303 \\
    & Misra et al & 0.338  & 0.280  & \textbf{0.396}  \\
    & BERT-Avg & 0.428  & 0.405  & 0.298  \\
    & BERT-MPro & \textbf{0.473}  & \textbf{0.451}  & 0.386  \\ \hline
WordNet & WNSP-noWSD & 0.114 & 0.045  & 0.266  \\
    & WNIC-noWSD & 0.152  & 0.128  & 0.318  \\
    & WNSP & 0.213  & 0.136  & 0.295  \\
    & WNIC & \textbf{0.467}  & \textbf{0.456}  & \textbf{0.448}  \\ \hline

Ensemble & BERT-MPro + WNIC & \textbf{0.547} & \textbf{0.531}  & \textbf{0.528}  \\ \hline
\end{tabular}
}

\caption{ \footnotesize Mean across categories (See Table~\ref{table:category-results} in Appendix~\ref{app:category} for results by category) of Spearman correlations for each method for the three datasets ($p < .001$). Largest correlations among singular methods for each dataset are bold-faced for BERT-based methods and WordNet-based methods. The last line is the overall best method and scores are also bold-faced. Note that for BERT-AVG, BERT-MPro, and BERT-SentEmb, the best performing layer is shown (layers 1, 10, and 11, respectively; see Appendices \ref{sec:sentemb_by_layer} and \ref{sec:avg_MP_hyper_perf} for a layer wise analysis).}
\label{table:all-results}

\centering
\small
\resizebox{3.7in}{!}{
\begin{tabular}{|l|ccc|c|c|} \hline
Category    & MPro & WNIC & MPro + WNIC & MPro vs WNIC & Avg. Increase\\\hline
Animals     & 0.665           & 0.522          & 0.697                & 0.524         & 0.125 \\
Birds       & 0.461           & 0.391          & 0.533                & 0.331        &  0.100 \\
Clothes     & 0.498          & 0.491          & 0.541                & 0.570         & 0.024 \\
Flowers     & 0.206         & 0.148          & 0.224                & 0.351         & 0.006 \\
Fruits      & 0.486           & 0.365          & 0.469                & 0.397       & 0.076  \\
Furniture   & 0.657          & 0.635          & 0.763                & 0.369        &  0.140 \\
Insects     & 0.366           & 0.429          & 0.408                & 0.476       & 0.043  \\
Instruments & 0.527           & 0.656          & 0.620                & 0.498       &  0.038 \\
Tools       & 0.366          & 0.584          & 0.607                & 0.274        &  0.134 \\
Vegetables  & 0.579         & 0.600          & 0.739                & 0.346         & 0.102 \\
Vehicles    & 0.389         & 0.314          & 0.418                & 0.529         & 0.053 \\ \hline
Mean            & 0.473           & 0.467          & 0.547           & 0.424    & 0.077 \\ \hline    
\end{tabular}}
\caption{Category wise Spearman correlations between each method and human rankings from (first three columns), Spearman correlations between BERT-MPro and WNIC rankings (fourth column), and the increase of performance of the ensemble method over MPro and WNIC (increases are averaged over the two individual methods). Results are shown for the Morrow and Duffy young adult dataset.}
\label{table:mpro vs wnic}

\end{table*}

\subsubsection{Polysemy Analysis}
\label{polysemy-results}
Experimental results show that disambiguation allows the improvements of BERT-MPro over BERT-Avg and WNIC over WNIC-noWSD. We now study whether the improvement is larger when the number of senses is larger. Our hypothesis is that the more polysemous a word, the more imprecise its representation without disambiguation, and the larger the performance increase from using disambiguation with the $maxsim$ operator. This analysis is difficult to do because: (i) we don’t have access to "true" senses, and (ii) both exemplar and category words can be polysemous. One solution is to use WordNet synsets as a proxy for senses. 

Looking at the estimated polysemy degree for the category words, we find that 8 of the 11 category words are associated with 2 senses or more, with the maximum ("tools") having 8 senses. Aligning with results for M\&D-young-adults, we find that BERT-MPro yields the largest correlation increases over BERT-Avg for the categories associated with polysemous category words: BERT-MPro yields above-average $\rho$ increase for 6 of the 8 polysemous categories, with the largest increase (+$0.2$) for the most ambiguous category word "tools". However, the same analysis on the same dataset with WNIC and WNIC-noWSD does not yield the same trend: while disambiguation leads to a greater relative increase, some of the category words with the least polysemy (animals, insects) have the largest performance improvement from noWSD to WSD (see Appendix \ref{polysemy_appendix} for more details). Looking at the average number of synsets for exemplars, it is not easy to see a trend in the performance improvements. Further analysis is needed and we leave a detailed study on polysemy and typicality for future work.

\section{Combining CLMs and Lexical Networks}
\label{sec:combining}

Given that WNIC and BERT-MPro rankings achieve the highest correlations, questions arise: how similar are the rankings produced by the two methods? If sufficiently different, how can we combine them? To answer these questions, we perform a series of analyses. We constrict this depth-first analysis to just the younger adults M\&D dataset.

\subsection{Method Complementarity Study}

BERT and WordNet provide different models of lexical meaning: BERT is word-oriented and exploits distributional statistical patterns, while WordNet is sense-oriented and exploits more abstract relations. To test whether this leads to different predictions, we compute category-wise Spearman correlations between BERT-MPro and WNIC rankings. The results are given in the fourth column in Table \ref{table:mpro vs wnic}. The values range from $0.274$ to $0.570$ with six categories below $0.4$. This moderate correlation shows some complementarity between the two rankings. Next, we present how different the two methods' rankings of certain exemplars can be in the first four columns of Table \ref{table:qual_mpro vs wnic}, showing how complementarity manifests on a granular level. This study raises the question of combining BERT-based methods with WordNet-based methods.

\subsection{A Simple Ensemble Method}
As an ensemble method, we take the raw scores of BERT-MPro and WNIC, convert them to $z$-scores (separately for each method and category), then sum them.\footnote{As an alternative ensemble method, we present a short study of KnowBERT, a CLM enhanced with WordNet knowledge, in Appendix \ref{knowbert}.} As shown in the last row of Table \ref{table:all-results}, BERT-MPro combined with WNIC achieves the highest correlation (greater than $0.5$) with human rankings across all datasets, closing the gap on the human performance given by the inter-dataset correlation between Rosch and M\&D older adults. The category Spearman correlations for BERT-MPro, WNIC and the ensemble are shown in Table \ref{table:mpro vs wnic}. We find that the lower the correlation between the two methods (second from right column), the larger the improvement by combining: we find a rank correlation of $-0.555$ between the inter-method correlation and the average correlation increase of the ensemble (the last column of Table \ref{table:mpro vs wnic}), showing that the more different the two methods rankings, the more ensembling increases performance. In Table~\ref{table:qual_mpro vs wnic} are rankings of certain words within their category, showing how the ensemble improves rankings for exemplars that are badly ranked by one of the methods.

\begin{table}[]
\centering
\small
\resizebox{\columnwidth}{!}{
\begin{tabular}{|ll|lll|l|} \hline
Category   & Word    & WNIC & MPro & WNIC + MPro & Human \\ \hline
Birds      & owl & 15    & 5        & 4       & 12    \\
Birds      & swallow & 5    & 22        & 11       & 11    \\
Clothes    & shirt    & 10   & 4        & 3       & 6   \\ 
Clothes    & suit    & 22   & 8         & 13       & 12   \\ 
Vegetables & lettuce & 7    & 16        & 11       & 11    \\
Vegetables & sprout & 17    & 11        & 12       & 12    \\\hline
\end{tabular}
}
\caption{Rankings of certain exemplars within their category, as scored by WNIC, BERT-MPro, WNIC + BERT-MPro, and the human ranking. Examples are from the Morrow and Duffy young adult dataset.}
\label{table:qual_mpro vs wnic}
\end{table}

\section{Conclusion}
\label{conclusion}
We show that BERT and WordNet-based methods are able to outperform previous methods in typicality prediction, but only with a disambiguation mechanism, either implicit through the selection of contexts or explicit via the estimation of distinct word senses. These results emphasize the importance of polysemy in assessing typicality, an issue that had been overlooked in previous work. We also show that BERT and WordNet provide complementary information that are relevant to modeling category structure: a simple ensemble of the two leads to the best correlations with human judgements.

We plan to further analyze the differences between CLMs and WordNet in typicality, and find more sophisticated ways to inject semantic knowledge into CLMs. Also, we want to use datasets in other languages, to find if the results generalize.

\section*{Limitations}

The human typicality judgements we use from \citet{morrow2005representation} is limited to English and uses only participants all from the UK, while the rankings from \citet{rosch1975cognitive} are all from the United States; thus, geographical/cultural biases could affect the typicality judgements (certain bird species or clothing exemplars could be more common in different areas, which could affect how human's judge their typicality). 

While we try to analyze exactly why disambiguation increases performance (BERT-MPro, WNIC-WSD) in Sec.~\ref{polysemy-results} and Appendix~\ref{polysemy-table}, we acknowledge this analysis is far from perfect: we don't have access to "gold" senses to measure polysemy (just linked WordNet synsets), and no single trend emerges from the analysis across all the methods. 

Lastly, one limitation in our BERT-based methods is that we use the \texttt{uncased} pretrained model. In hindsight, we might have avoided some ambiguous word uses by using a \texttt{cased} model instead: for example, the model might have been able to more easily distinguish between "Apple", the company, and "apple", the fruit. However, the impact might have been marginal, as the model can still rely on contextual clues of the surrounding text, and such cases of ambiguity are arguably rare in the typicality datasets. Another possible limitation, specific to the BERT-MLM and BERT-SentEmb methods, comes from the current sentence sampling procedure, as it is based on the category word only, thus possibly introducing noise for words that have homonyms with different POS (e.g., noun-verb honomyms).  

\section*{Acknowledgements}
We would like to thank the three anonymous EACL reviewers for their helpful comments on this paper. This research was funded by Inria Exploratory Action COMANCHE, as well as by the joint IMPRESS project between Inria and DFKI.

\bibliographystyle{acl_natbib}
\bibliography{custom}

\appendix

\section{Appendix}

\subsection{Dataset Preprocessing}
\label{dataset:preprocessing}

We follow the preprocessing procedure of \citet{heyman2019can}, discarding multi-word examples (such as \textit{red cabbage}) and examples containing punctuation. Although these cases could have been handled by BERT-based methods using subtokens, the composition of several terms is non-trivial for word embedding based models \citep{lenci2018distributional}, and we are more concerned with comparing category structure than how the different models handle multi-word examples or punctuation. Also following \citet{heyman2019can}, we discard categories with less than $20$ examples leaving $11$ categories in the \citet{morrow2005representation} datasets and 10 in the \citet{rosch1975cognitive} dataset. As some of the baseline models have fixed vocabularies (PPMI-SVD, Word2Vec, WordNet), we remove any out of vocabulary examples from the dataset. While we could have used median or minimum typicality scores for out of vocabulary examples, we wanted to ensure that the comparison of performance was fair across all methods by only evaluating on examples that have a valid typicality score for all methods.

\begin{table}[]
\centering
\small

\begin{tabular}{|l|l|l|} \hline
Category    & M\&D  & Rosch \\ \hline
Animals     & 128    &             \\
Birds       & 73     & 54             \\
Clothes     & 79    & 48              \\
Flowers     & 45    &               \\
Fruits      & 59    & 44              \\
Furniture   & 39    & 45              \\
Insects     & 44    &               \\
Instruments & 59    &               \\
Sports & & 47 \\
Tools       & 59    & 49              \\
Toys & & 42 \\ 
Vegetables  & 29    & 42              \\
Vehicles    & 68    & 46               \\ 
Weapons & &  52\\\hline
Total & 682 & 470 \\ \hline
\end{tabular} 
\caption{\footnotesize Number of exemplars per category for two datasets: \citet{morrow2005representation} and \citet{rosch1975cognitive}.}
\label{table:dataset}
\end{table}

\subsection{Dataset Agreement}
\label{dataset_agreement}
Tables \ref{table:ya_oa_spearman} and \ref{table:dataset_intersection_spearman} show the category-wise agreement between the human rankings of the three datasets. Notice that the agreement between the two Morrow and Duffy datasets are much higher than that of Rosch and the two M\&D datasets.

\begin{table}[]
\centering
\small
\begin{tabular}{|l|l|} \hline
Category            & Spearman \\ \hline
Animals     & 0.916                      \\
Birds       & 0.868                      \\
Clothes     & 0.858                      \\
Flowers     & 0.747                      \\
Fruits      & 0.891                      \\
Furniture   & 0.778                      \\
Insects     & 0.897                      \\
Instruments & 0.909                      \\
Tools       & 0.791                      \\
Vegetables  & 0.885                      \\
Vehicles    & 0.887                      \\ \hline
Mean        & 0.857                     \\ \hline
\end{tabular}
\caption{Spearman correlations between young adult and older adult typicality rankings for \citet{morrow2005representation}.}
\label{table:ya_oa_spearman}
\end{table}

\begin{table}[]
\centering
\small
\begin{tabular}{|l|l|l|l|} \hline
           & Young vs Rosch & Old vs Rosch & N  \\ \hline
Birds      & 0.561                       & 0.449              & 43 \\
Clothes    & 0.736                       & 0.647              & 33 \\
Fruits     & 0.871                       & 0.680              & 38 \\
Furniture  & 0.762                       & 0.547              & 20 \\
Tools      & 0.535                       & 0.613              & 25 \\
Vegetables & 0.326                       & 0.227              & 22 \\
Vehicles   & 0.802                       & 0.852              & 29 \\ \hline
Mean       & 0.656                       & 0.574              & 30 \\ \hline
\end{tabular}
\caption{Spearman correlations between the young adult and older adult rankings of \citet{morrow2005representation} and \citet{rosch1975cognitive} on the intersection of categories and exemplars between the two datasets. The size of the intersection is shown on the far right column.}
\label{table:dataset_intersection_spearman}
\end{table}

\subsection{Sentence Sampling}
\label{dataset:sent_sampling}
For an auxiliary textual dataset $S$, we use Wikipedia, specifically searching for sentences that contain the singular form of each category word or exemplar, in order to have comparable contexts. We found text from Wikipedia dumps can be noisy (such as: summarization tables, lists of related topics, other sequences that are not actual sentences), so we remove sentences that are too short or long (number of words must be between 5 and 200). For BERT-AVG and BERT-MPro, we sampled 300 sentences for every exemplar and category word. For BERT-MLM and BERT-SentEmb, we sampled 10000 sentences for each category.

\subsection{PPMI-SVD and Word2Vec Details}
\label{baseline_details}
The PPMI matrix was computed from English Wikipedia using a window size of $2$. We experimented with a window size of $5$, but found the performance to be worse. We also use off-the-shelf Word2Vec \citep{mikolov2013distributed} embeddings (skip-gram with negative sampling, 300 dimensions, GoogleNews training corpus).

\subsection{Runtimes}
\label{runtimes}
Frequency, PPMI-SVD, Word2vec, and both WordNet baseline methods take <5 seconds to compute typicality scores on CPU. The Wikipedia word counts used in Frequency and WordNet-IC take ~1hr to compute on CPU. The sentence sampling from Wikipedia takes ~2hrs to complete. BERT-WordPiece takes ~1 min to complete on CPU. BERT-AVG and BERT-MLM take ~3hrs to complete on the GPU mentioned in the text; BERT-SentEmb, ~5hr; and BERT-MPro, ~10hr.

\subsection{Results by Category}
\label{app:category}
Table \ref{table:category-results} shows the Spearman's correlation values for each individual category. Note that just the top performing methods are shown.

\begin{table*}[h]

\centering
\footnotesize
\resizebox{440pt}{!}{
\begin{tabular}{|l|p{0.7cm}p{0.7cm}p{0.7cm}|p{0.7cm}p{0.7cm}p{0.8cm}|p{0.7cm}p{0.7cm}p{0.8cm}|p{0.7cm}p{0.7cm}p{0.7cm}|p{0.7cm}p{0.7cm}p{0.8cm}|} \hline
            & \multicolumn{3}{l|}{Misra et al} & \multicolumn{3}{l|}{Bert-MLM-Taxo} & \multicolumn{3}{l|}{Bert-Avg} & \multicolumn{3}{l|}{Bert-MPro}               & \multicolumn{3}{l|}{WNIC}        \\
Category    & M\&D-Y    & M\&D-O    & Rosch    & M\&D-Y     & M\&D-O    & Rosch     & M\&D-Y   & M\&D-O   & Rosch   & M\&D-Y & M\&D-O & Rosch & M\&D-Y   & M\&D-O  & Rosch  \\ \hline
Animals     & 0.558     & 0.525     &          & 0.528      & 0.508     &           & 0.569    & 0.530    &         & 0.665  & 0.649  &    & 0.522 & 0.555 &        \\
Birds       & 0.284     & 0.130     & 0.308    & 0.442      & 0.332     & 0.191     & 0.440    & 0.311    & 0.203   & 0.461  & 0.317  & 0.169                      & 0.391                                             & 0.310                      & -0.014 \\
Clothes     & 0.205     & 0.115     & 0.461    & 0.527      & 0.587     & 0.171     & 0.506    & 0.528    & 0.301   & 0.498  & 0.573  & 0.484                      & 0.491                                             & 0.487                      & 0.570  \\
Fruits      & 0.232     & 0.191     & 0.354    & 0.287      & 0.357     & 0.278     & 0.390    & 0.338    & 0.406   & 0.486  & 0.433  & 0.592                      & 0.365                                             & 0.403                      & 0.442  \\
Furniture   & 0.600     & 0.440     & 0.548    & 0.577      & 0.429     & 0.484     & 0.629    & 0.598    & 0.320   & 0.657  & 0.602  & 0.235                      & 0.635                                             & 0.553                      & 0.658  \\
Flowers     & 0.240     & 0.146     &          & 0.122      & 0.068     &           & 0.152    & 0.011    &         & 0.206  & 0.111  &    & 0.148 & 0.105 &        \\
Insects     & 0.175     & 0.328     &          & 0.411      & 0.379     &           & 0.437    & 0.473    &         & 0.366  & 0.503  &                            & 0.429                                             & 0.415                      &        \\
Instruments & 0.309     & 0.331     &          & 0.667      & 0.715     &           & 0.626    & 0.686    &         & 0.527  & 0.602  &                            & 0.656                                             & 0.659                      &        \\
Sports      &           &           & 0.467    &            &           & 0.536     &          &          & 0.530   &        &        & 0.379                      &                                                   &                            & 0.498  \\
Tools       & 0.421     & 0.259     & 0.100    & 0.066      & 0.220     & 0.220     & 0.160    & 0.257    & 0.040   & 0.366  & 0.318  & 0.086                      & 0.584                                             & 0.486                      & 0.494  \\
Toys        &           &           & 0.143    &            &           & -0.001    &          &          & -0.089  &        &        & 0.398                      &                                                   &                            & 0.155  \\ 
Vegetables  & 0.390     & 0.312     & 0.241    & 0.364      & 0.281     & -0.006    & 0.501    & 0.397    & 0.073   & 0.579  & 0.482  & 0.088                      & 0.600                                             & 0.646                      & 0.409  \\
Vehicles    & 0.307     & 0.301     & 0.682    & 0.364      & 0.444     & 0.733     & 0.293    & 0.323    & 0.743   & 0.389  & 0.373  & 0.651                      & 0.314                                             & 0.396                      & 0.701  \\
Weapons     &           &           & 0.655    &            &           & 0.429     &          &          & 0.454   &        &        & 0.779                      &                                                   &                            & 0.567  \\\hline

\end{tabular}
}
\caption{Category wise Spearman correlations with human typicality rankings for the four best BERT-based models and the best WordNet-based model.}
\label{table:category-results}
\end{table*}

\subsection{BERT-SentEmb Performance by Layer}
\label{sec:sentemb_by_layer}

    \includegraphics[width=8.0cm]{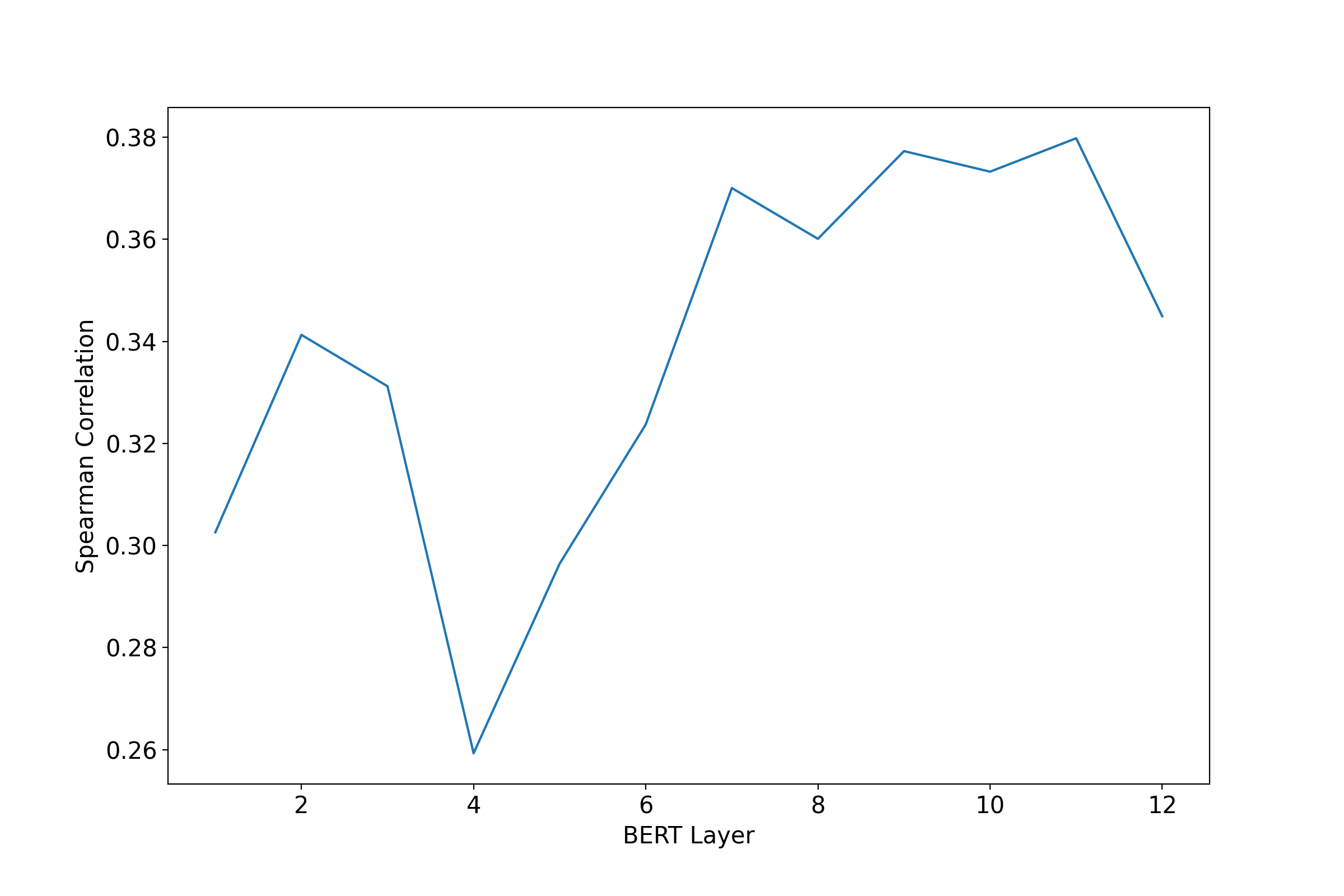}
    \label{fig:sent_bert_layer}
The above figure shows the mean Spearman correlations by layer for the BERT NSP method.

\subsection{BERT-AVG and MultiPrototype Hyperparameter Performance}
\label{sec:avg_MP_hyper_perf}

\includegraphics[width=8.0cm]{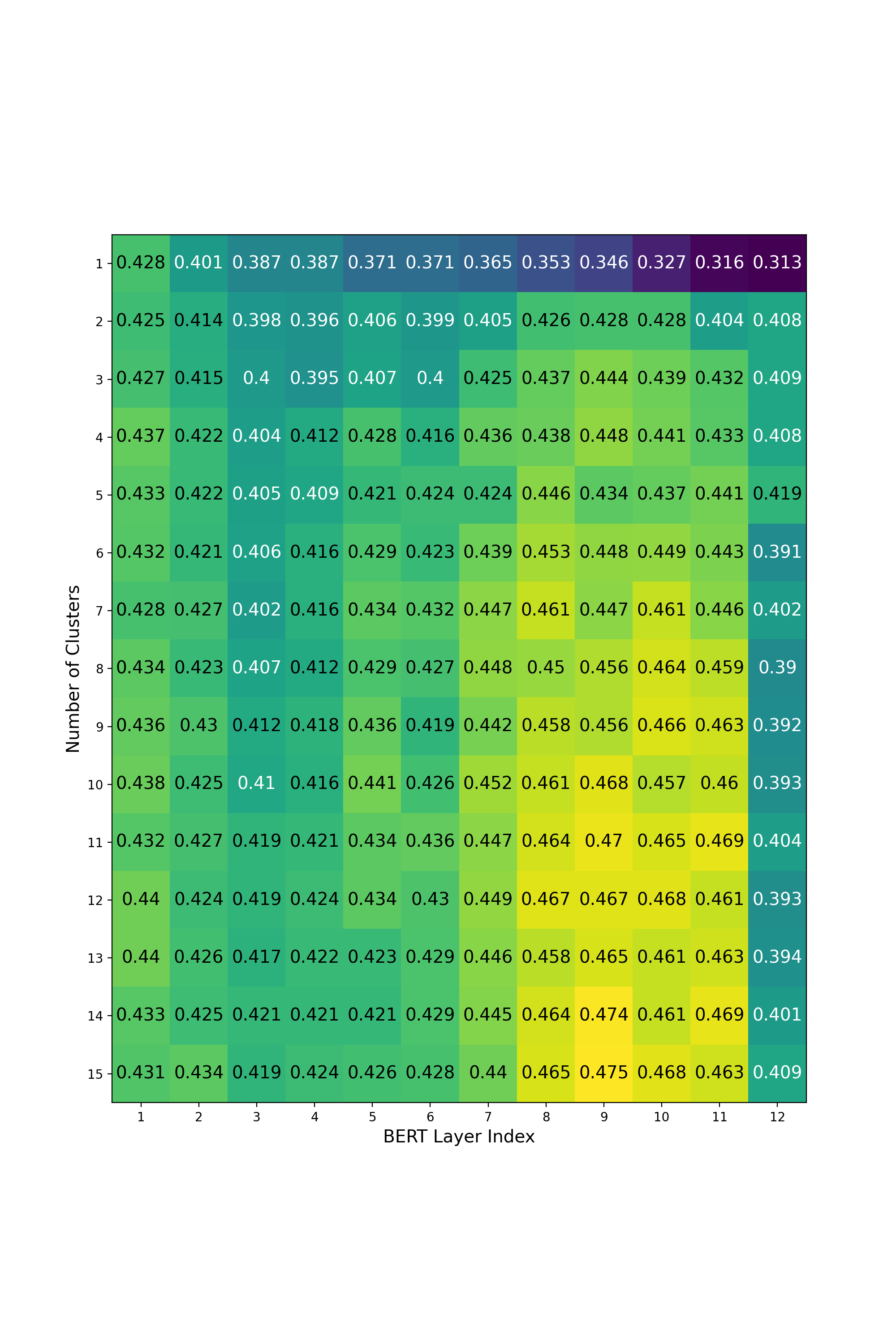}

\label{fig:bert_heatmap_full}
The above image shows average Spearman's correlation across all categories for BERT multi-prototype embeddings for each combination of layer index and number of cluster. Note that the top row (number of clusters = 1) is the same as the simple average of contextual representations.

\subsection{Detailed Polysemy Analysis}
\label{polysemy_appendix}
To see whether the $maxsim$ disambiguation mechanism is actually increasing performance more for more polysemous examples, we can compare the performance of the same methods with and without disambiguation when evaluated on words with different number of linked WordNet synsets (a proxy for the degree of polysemy). As shown in Table \ref{polysemy-table}, in general, BERT-MPro's disambiguation increases performance over BERT-Avg more for more polysemous categories (as measured by the synsets linked to the category word). However, this general trend is not seen when comparing WNIC to WNIC-noWSD.

\begin{table*}[h]
\centering
\small
\begin{tabular}{|l|ll|ll|}
\hline
Category    & MPro inc. & WNIC inc. & Cat. Synsets  & Exem. Synsets       \\  \hline
Animals     & 0.096         & 0.517             & 1  & 2.992      \\
Birds       & 0.021         & 0.362             & 6  &  2.712          \\
Clothes     & -0.009        & 0.397             & 4  &  4.241            \\
Flowers     & 0.055         & 0.028             & 4  &  2.022            \\
Fruits      & 0.096         & 0.409             & 5  & 3.102            \\
Furniture   & 0.028         & 0.165             & 1  & 4.590       \\
Insects     & -0.071        & 0.432             & 2  & 2.705         \\
Instruments & -0.099        & 0.456             & 1  &  2.423            \\ 
Tools       & 0.206         & 0.155             & 8  & 5.136          \\
Vegetables  & 0.078         & 0.317             & 2   & 3.345           \\
Vehicles    & 0.095         & 0.222             & 4     & 3.412         \\ \hline
\end{tabular}
\caption{Increase in Spearman's correlation from using BERT-MPro over BERT-Avg (middle left column) and from using WNIC over WNIC-noWSD (middle right column) for each category. The number of linked WordNet synsets for each category and their exemplars are shown in the right columns. Results are shown for the \citet{morrow2005representation} young adult dataset.}
\label{polysemy-table}
\end{table*}

\subsection{KnowBERT Results}
\label{knowbert}
The combination of CLMs and knowledge graphs is an active research area, mainly the design of knowledge enhanced CLMs (i.e. \citet{kadapter}, \citet{knowledgeaware}). We use KnowBERT \cite{knowbert}, in which knowledge bases are embedded into BERT via an integrated entity linker, which retrieves relevant entity embeddings and updates the hidden states (we leave an in-depth description to the paper). We consider a KnowBERT model with WordNet as the knowledge base\footnote{Candidate WordNet synsets are found for KnowBERT using a similar procedure as we use to find all possible synsets.} with the multi-prototype method for computing typicality scores. We get an average Spearman correlation of 0.446, below that of WNIC and BERT-MPro, showing that a general method for enhancing CLMs with WordNet does not work well for typicality.

\end{document}